\documentclass{llncs}
\usepackage{lmodern}
\usepackage{hyperref}

\usepackage{subfig, graphicx}
\usepackage{algorithm}
\usepackage{listings}
\usepackage{fancyvrb}

\newcommand{\prop}[1]{\textit{#1}}
\newcommand{\ocls}[1]{\textit{#1}}
\newcommand{\ANNConfiguration}{\ocls{ANNConfiguration}}
\newcommand{\Network}{\ocls{Network}}
\newcommand{\TaskCharacterization}{\ocls{TaskCharacterization}}
\newcommand{\DataCharacterization}{\ocls{DataCharacterization}}

\newcommand{\Layer}{\ocls{Layer}}
\newcommand{\ActivationLayer}{\ocls{ActivationLayer}}
\newcommand{\AggregationLayer}{\ocls{AggregationLayer}}
\newcommand{\SeparationLayer}{\ocls{SeparationLayer}}
\newcommand{\ModificationLayer}{\ocls{ModificationLayer}}
\newcommand{\InOutLayer}{\ocls{InOutLayer}}
\newcommand{\HiddenLayer}{\ocls{HiddenLayer}}

\newcommand{\InputLayer}{\ocls{InputLayer}}
\newcommand{\OutputLayer}{\ocls{OutputLayer}}

\newcommand{\TrainingStrategy}{\ocls{TrainingStrategy}}
\newcommand{\TrainingSession}{\ocls{TrainingSession}}
\newcommand{\TrainingStep}{\ocls{TrainingStep}}
\newcommand{\TrainingLoop}{\ocls{TrainingLoop}}

\newcommand{\Function}{\ocls{Function}}
\newcommand{\ActivationFunction}{\ocls{ActivationFunction}}
\newcommand{\ObjectiveFunction}{\ocls{ObjectiveFunction}}
\newcommand{\Metric}{\ocls{Metric}}

\newcommand{\Dataset}{\ocls{Dataset}}
\newcommand{\Labelset}{\ocls{Labelset}}

\newcommand{\DatasetPipe}{\ocls{DatasetPipe}}
\newcommand{\TrainedModel}{\ocls{TrainedModel}}

\newcommand{\NetworkEvaluation}{\ocls{NetworkEvaluation}}

\newcommand{\ie}{i.e.}
\newcommand{\eg}{e.g.}

\newcommand{\name}{ANNETT-O}

\usepackage[backgroundcolor=yellow!40,bordercolor=red,linecolor=red,textsize=footnotesize]{todonotes}

\title{\name{}}
\subtitle{An Ontology for Describing Artificial Neural Network Evaluation, Topology and Training}
\titlerunning{\name{} Ontology}

\author{Iraklis A. Klampanos \and Athanasios Davvetas \and Antonis Koukourikos \and Vangelis Karkaletsis}
\authorrunning{Klampanos et. al.}
\institute{National Centre for Scientific Research ``Demokritos'', \\ Agia Paraskevi, Greece \\\email{iaklampanos@iit.demokritos.gr}
}

\begin{document}
\maketitle

\begin{abstract}
Deep learning models, while effective and versatile, are becoming increasingly complex, often including multiple overlapping networks of arbitrary depths, multiple objectives and non-intuitive training methodologies. This makes it increasingly difficult for researchers and practitioners to design, train and understand them. In this paper we present \name{}, a much-needed, generic and computer-actionable vocabulary for researchers and practitioners to describe their deep learning configurations, training procedures and experiments. The proposed ontology focuses on topological, training and evaluation aspects of complex deep neural configurations, while keeping peripheral entities more succinct. Knowledge bases implementing \name{} can support a wide variety of queries, providing relevant insights to users. In addition to a detailed description of the ontology, we demonstrate its suitability to the task via a number of hypothetical use-cases of increasing complexity.
\end{abstract}

\section{Introduction}
\label{sec:intro}
Deep learning is a machine learning discipline that focuses on the specification and training of deep neural networks. Deep learning is currently the driving force behind a multitude of AI applications, such as speech recognition, computer vision, robotics and others \cite{Bengio2009,Lecun2015}, with applications as diverse as the classification of exoplanets \cite{Shallue2017}, inverse atmospheric dispersion \cite{Klampanos2017}, weather forecasting \cite{Ghaderi2017} and information retrieval \cite{Mitra2017}.

Data scientists, engineers and machine learning practitioners often tackle diverse problems via the use of increasingly complex configurations of deep artificial neural networks (ANNs). Such complex configurations comprise series of co-trained networks, where each network's objective directly or indirectly affects the objectives of its sibling networks. This interplay contributes to the overall objective of the deep learning configuration. Examples of such multi-network configurations are the generative adversarial network \cite{NIPS2014_5423}, the adversarial autoencoder \cite{Makhzani2015}, and others.

Despite their effectiveness, working with deep neural networks has a number of shortcomings. Arguably the most important one is the difficulty for users to understand and argue about their internal operation, hence why they often describe them as ``black boxes''. Another disadvantage is that they can be difficult to train effectively, often requiring multiple trial-and-error iterations. Topological choices, such as depth and connectivity, as well as the choice of hyperparameters, are also often the result of trial-and-error. These difficulties become greater as the complexity of neural network configurations increases.

In this paper we present \name{}, an ontology which can capture and link many of the topological, training and evaluation characteristics of existing and in-development ANN configurations, in order to create knowledge bases that could drive the design of deep learning solutions. As complexity increases, ubiquitous deep learning can only be maintained if the knowledge and intuition gained can be harvested and encoded in useful ways. The Semantic Web \cite{Berners-Lee2001}, and the ontology specification language OWL\footnote{\url{https://www.w3.org/OWL/}} can enable the creation of abstractions and tools towards this direction. The expressiveness and extensibility of OWL and the distributed nature of the Semantic Web are important prerequisites for creating useful knowledge bases and resources for users.

This paper provides a generic-enough, usable and computer-actionable topology for researchers and practitioners to describe their deep learning configurations, training procedures and experiments. The proposed ontology connects three aspects of deep learning design:
\begin{description}
    \item[Topology] The connectivity of complex, multi-network, configurations;
    \item[Training] The algorithms used for training complex ANN configurations;
    \item[Evaluation] Quantitative performance information.
\end{description}

In the next section we present relevant work. In Section \ref{sec:classes} we present the main classes of the ontology for describing ANN configuration evaluation, topology and training. In Section \ref{sec:use-cases} we present three exemplary use-cases of increasing complexity, while in Section \ref{sec:queries} we provide sample queries in SPARQL. In Section \ref{sec:conclusions} we provide pointers for future work.

\section{Related Work}
\label{sec:related}

To the best of our knowledge, \name{} is the first ontology suitable to describe complex, multi-network neural configurations with an emphasis on studying, understanding and improving future algorithms. The closest ontology to \name{} is the Artificial Neural Network Ontology -- ANNO\footnote{\url{https://tw.rpi.edu/web/Courses/Ontologies/2016/projects/ArtificialNeuralNetworkOntology}}, whose primary purpose, according to its authors, is to recommend weight initialization for Keras\footnote{\url{https://keras.io/}, viewed March 2018} neural network models. Despite its seemingly narrow scope, ANNO includes many generally useful concepts, such as classes for algorithms and functions. \name{} follows a similar class arrangement for the common concepts, however it has an entirely different focus.

Another relevant resource is the Predictive Model Markup Language -- PMML \cite{Grossman1999}. PMML is a detailed and mature XML schema for describing predictive models, including neural networks. PMML is in use in many applications, such as the popular KNIME data analytics platform\footnote{\url{https://www.knime.com/}, viewed March 2018} \cite{knime,Morent2011}, in manufacturing \cite{Lee2014}, and elsewhere. One of the core objectives of PMML is to allow for cross-platform execution of pre-trained predictive models. This requires a high level of detail regarding the numerical aspect of the models (\eg{} the learned weights in an ANN) while it focuses less on the methodology applied to training them. In addition, PMML being an XML schema makes it less flexible and extensible than it would be required in use-cases addressed by \name{}. However, aspects of PMML could potentially complement \name{} in certain application contexts, but this is left as future work.

\section{The Ontology}
\label{sec:classes}
\begin{figure}[t]
    \centering
    \includegraphics[width=\textwidth]{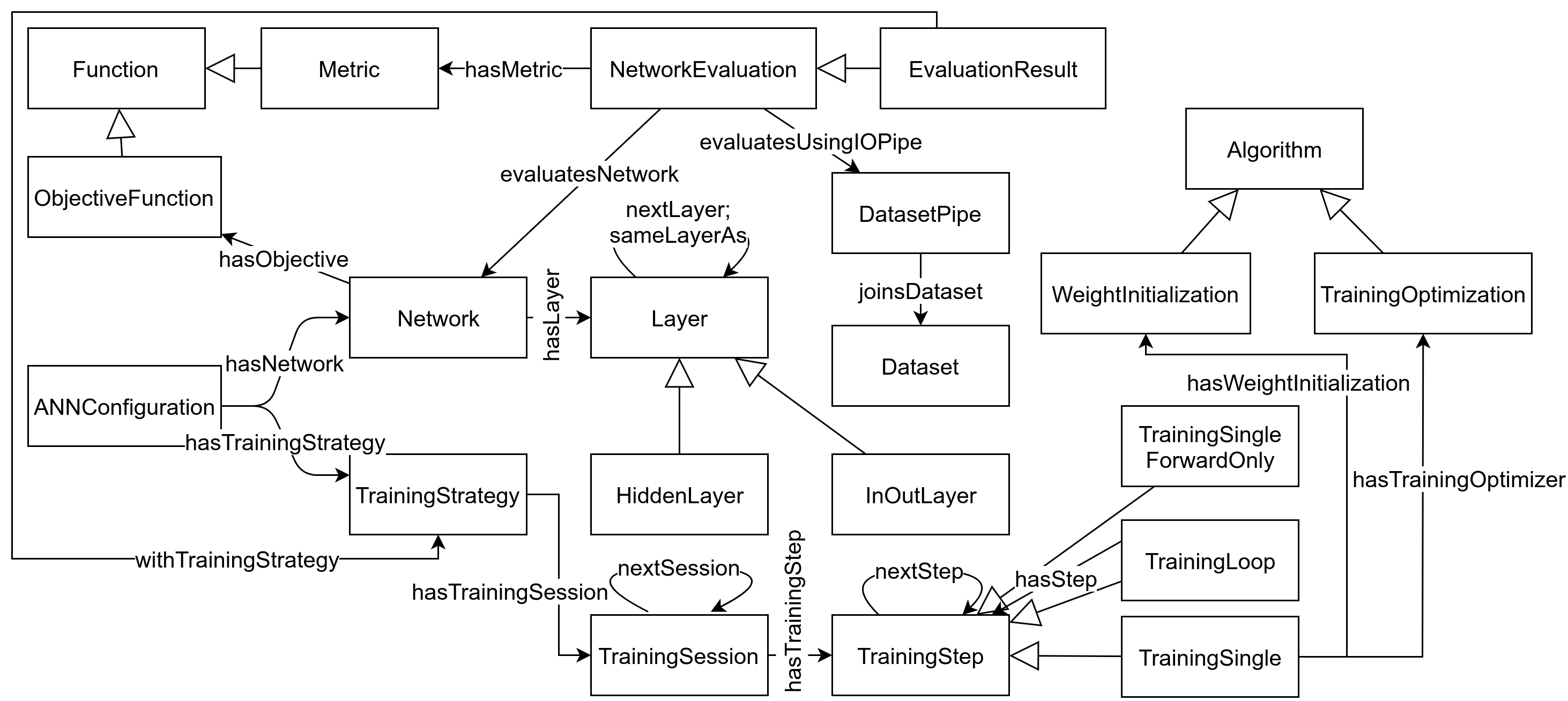}
    \caption{The main classes and object properties of the \name{} ontology. Triangular arrowheads denote subclass-of relations.}
    \label{fig:main-classes}
\end{figure}

Figure \ref{fig:main-classes} shows the main classes and interactions of the ontology. \name{} defines 160 classes, 50 object properties and 32 data properties. 
The ontology is assigned the permanent identifier \url{http://w3id.org/annett-o/}, also acting as a URL redirecting to the resource's current location: \url{https://github.com/iaklampanos/annett-o}. 
\name{} is licensed under a Creative Commons Attribution 4.0 International License\footnote{\url{https://creativecommons.org/licenses/by/4.0/}}.


\name{} is designed with multi-network, multi-objective configurations in mind. Such configurations are expressed through the instantiation of the \ANNConfiguration{} class, which relates neural network individuals as part of an overall configuration. In the simplest case of describing a single-objective neural network, an \ANNConfiguration{} individual is associated with a single \Network{} individual. Each neural network can be described in terms of its layers, modeled via the \Layer{} class or one of its subclasses. Layer subclasses allow for the description of different types of ANNs, such as feedforward and recurrent configurations. \name{} does not  model networks down to the node (or neuron) level since it is common practice that each layer has a uniform behavior (activation or other transformation) implemented by its constituent nodes.

Multi-network configurations involve multiple objectives and complex training methodologies. Often constituent networks get trained one after another per batch iteration\footnote{ANNs are typically trained iteratively in data batches, which are small subsets of the complete dataset.},  while in other cases one or more networks might get pre-trained followed by a more complete ``fine-tuning'' procedure. In more involved training procedures there may be multiple training iterations of a subset of networks before training another subset, within the same batch iteration. The link between a training sequence and the choice of layers and their connectivity is often unclear. \name{} enables researchers to study and extract insights from both, therefore discovering best practices and improving their algorithms.

\TrainingStrategy{} is the main training class. A training strategy is composed of a series of \TrainingSession{}s, with each session interpreted as a complete training procedure over an entire training dataset. Each training session defines at least one \TrainingStep{}, which denotes the training within batch iterations. Training steps may also form a sequence with each step denoting either the training of a single network, a pass through a network creating a new dataset to be used in a subsequent training session, or a loop of steps.
   
The choices for topology and for training strategies affect the performance of an ANN configuration. \name{} supports describing network evaluation results via its \NetworkEvaluation{} class, which is connected to both topological choices and to training strategies.

\name{} allows for the description and reasoning of topological, training and evaluation characteristics of  complex, multi-ANN configurations. In the remainder of this section we introduce the main classes involved.

\subsection{Topology}
\label{sec:topology}
In \name{} the topology of an ANN configuration is described via the description of its constituent neural networks, their corresponding layers and the way they connect, \ie{} their activation paths.

\vspace{.4em}
\noindent
\textbf{\ANNConfiguration{}}
A neural network configuration potentially comprises multiple networks. A neural network configuration has a well-defined purpose, while the networks constituting the array may be disjoined with one another. An example of such an ANN configuration may require a set of networks to be trained separately from another set, while collectively having a single purpose or a common, in the loose sense, objective.
%

\vspace{.4em}
\noindent
\textbf{\Network{}}
Individuals of the \Network{} class describe neural networks. Each network must be associated with at least one ANN configuration. Networks are described in terms of connected layers. A network may share certain layers with other networks. In \name{} this is implemented via the object property \prop{sameLayerAs}, discussed below. Each \Network{} individual can only have a single objective, described by a single objective function.

\vspace{.4em}
\noindent
\textbf{\Layer{}}
The class \Layer{} and its subclasses describe various types of layer that may be present in a neural network. Immediate subclasses include \HiddenLayer{} and \InOutLayer{}. \HiddenLayer{} has a number of subclasses, namely \ActivationLayer{}, \AggregationLayer{}, \SeparationLayer{} and \ModificationLayer{}. \ActivationLayer{} individuals describe  layers with trainable weights carrying activation functions. Activation layers are associated to activation functions via the object property \prop{hasActivationFunction}. Modification layers modify their inputs in a static, non-trainable way, \eg{} by introducing a fixed amount of noise. The rest are layers with special roles in the network topology, \eg{} a separation layer may denote the cloning of the previous layer's outputs into multiple following layers.

\vspace{.4em}
\noindent
\textit{Connecting layers:} Each network is associated with a number of layers via its object property \prop{hasLayer}. The connectivity of layers within networks is described via the object properties \prop{nextLayer} and its symmetric property \prop{previousLayer}. Each \Layer{} individual can connect to at most one layer following or coming before it, with the following exceptions: \InputLayer{}  may not have layers connecting into it; \OutputLayer{}  may not have layers connecting from it; \SeparationLayer{}  may have more than one layers coming after it; and \AggregationLayer{}  may have more than one layers connecting into it.

In \name{} each \Layer{} individual can be associated strictly with a single \Network{} individual. This restriction allows the ontology to enforce the cardinality rules regarding \prop{nextLayer} and \prop{previousLayer} described above without increasing complexity -- if \Layer{} individuals belonged to more than one networks, \prop{nextLayer} would need to represent a three-party relation between one \Network{} and two \Layer{} individuals. Even though there are patterns to describe n-ary relations\footnote{\url{https://www.w3.org/TR/swbp-n-aryRelations/}, viewed March 2018}, it was decided that this would increase the complexity of the ontology without users gaining on flexibility or descriptiveness. \name{} describes the presence of common layers in different networks by introducing the \prop{sameLayerAs} object property between two \Layer{} individuals.

\subsection{Training}
\label{sec:training}

\vspace{.4em}
\noindent
\textbf{\TrainingStrategy{}}
A \TrainingStrategy{} individual describes the steps taken to train a complex ANN configuration. This may involve one or more sequential \TrainingSession{}s pointed at via the \prop{hasTrainingSession} object property.

\vspace{.4em}
\noindent
\textbf{\TrainingSession{}}
Each \TrainingSession{} individual represents a complete training session, \ie{} a complete training process over a dataset. The dataset is expected to be used in batches, per standard practice. For instance, if the training of an ANN configuration does not depend on prior training of one of its constituent networks then a single \TrainingSession{} suffices. On the other hand, the case of a network needing to be pre-trained on a dataset before it can be fine-tuned, would require the use of two \TrainingSession{} individuals. A chain of training sessions can be formed via the \prop{nextTrainingSession} object property, which may appear at most once per \TrainingSession{} individual.

\vspace{.4em}
\noindent
\textbf{\TrainingStep{}}
Each training session is composed of a sequence of \TrainingStep{}s implemented by the property \prop{nextTrainingStep}. The sequence of training steps is expected to be repeated for each batch of the training session dataset(s). Each training step is typically associated with a neural network, the sequence therefore allowing to model the simultaneous training of neural networks\footnote{Simultaneous training refers to the alternate training of two or more networks within the same batch iteration.} of the ANN configuration -- see Section \ref{sec:use-cases} for examples.

During simultaneous training, sometimes a network is trained for a number of times, or until a condition on its performance has been satisfied, before other networks are trained. This is accomplished by the \TrainingLoop{} class, which is a subclass of \TrainingStep{} and can therefore be part of a training step sequence. A training loop itself contains a sequence of training steps performed for a number of repetitions or until a condition has been met. Using \TrainingLoop{} individuals allows users to model training session such as ``train network $A$ for 5 times and then train network $B$, before proceeding to the next training iteration''.

\subsection{Evaluation}
In \name{} evaluation results are modeled via the \NetworkEvaluation{} class. A \NetworkEvaluation{} individual associates \Network{}, \ANNConfiguration{}, \TrainingStrategy{}, \Metric{} and \Dataset{} individuals. The dataset represents a training dataset, while the metric is the evaluation metric used. The evaluation score is recorded via the data property \prop{eval\_score}. \NetworkEvaluation{} may optionally record a timestamp for the evaluation via the data property \prop{eval\_date}. It follows that an ANN configuration may involve multiple evaluations involving different neural networks, using different metrics and datasets, for various training strategies.

\subsection{Auxiliary Classes}

\textbf{\Function} The \Function{} class and its subclasses describe reusable mathematical functions and are associated with a number of the \name{} concepts, \eg{} activation layers, introduced below, are associated with \ActivationFunction{}s via the \prop{hasActivationFunction} property. Function individuals may provide the function's mathematical form in some pre-agreed notation (\eg{}, in \LaTeX{}) using the \prop{function\_math} data property.

\vspace{.4em}
\noindent 
\textbf{\Dataset} Individuals of this class describe a homogeneous dataset used for training and evaluation purposes. A dataset may denote a well-known resource, such as MNIST\footnote{\url{http://yann.lecun.com/exdb/mnist/}, viewed March 2018}, a private or application-specific dataset, or a transient dataset. For instance, transient or temporary datasets may be created to aid during the training of an ANN configuration. A \Labelset{} (a subclass of \Dataset{})
may be used to denote a set of labels useful in classification tasks.

\vspace{.4em}
\noindent
\textbf{\DatasetPipe{}} Individuals of this class associate \InOutLayer{} to \Dataset{} individuals. This allows for the description of multiple connections to and from datasets in a number of cases, such as in the definition of an evaluation strategy.

\vspace{.4em}
\noindent
\textbf{\TrainedModel} Individuals of this class describe a trained model, namely the result of training taking place over a complete neural network configuration. This information may be useful for studying distributions of weights or for replicating ANN configurations in different contexts. Especially for the latter, PMML \cite{Grossman1999} descriptions may also be linked via subclassing.

\vspace{.4em}
\noindent
\textbf{\TaskCharacterization} Individuals of this class and its subclasses may be used to characterize  the purpose of a neural \Network{}. For instance, \TaskCharacterization{} subclasses include \ocls{Clustering}, \ocls{Classification}, \ocls{Generation}, etc.

\vspace{.4em}
\noindent
\textbf{\DataCharacterization} Individuals of this class and its subclasses can be used to characterize  a \Dataset{}. For instance, \TaskCharacterization{} subclasses include \ocls{NumericanDigits}, \ocls{Flora}, \ocls{People}, etc.

\section{Example Use-Cases}
\label{sec:use-cases}

In this section we present three exemplary \name{} use-cases of increasing complexity: a simple ANN used in classification tasks \cite[Ch. 6]{Goodfellow2016}, a generative adversarial network (GAN) \cite{NIPS2014_5423}, and an adversarial autoencoder (AAE) \cite{Makhzani2015}. These examples are distributed along with the ontology.

\begin{figure}
    \centering
    \caption{Example ANN configurations of increasing complexity. Layers are labeled to allow comparisons to the \name{} representations of Figures \ref{fig:GAN-individual-nets} and \ref{fig:AAE-individual-nets}.}
    \subfloat[Simple classification network. In name{} it can be described as a single \Network{}.]{\includegraphics[width=.42\textwidth]{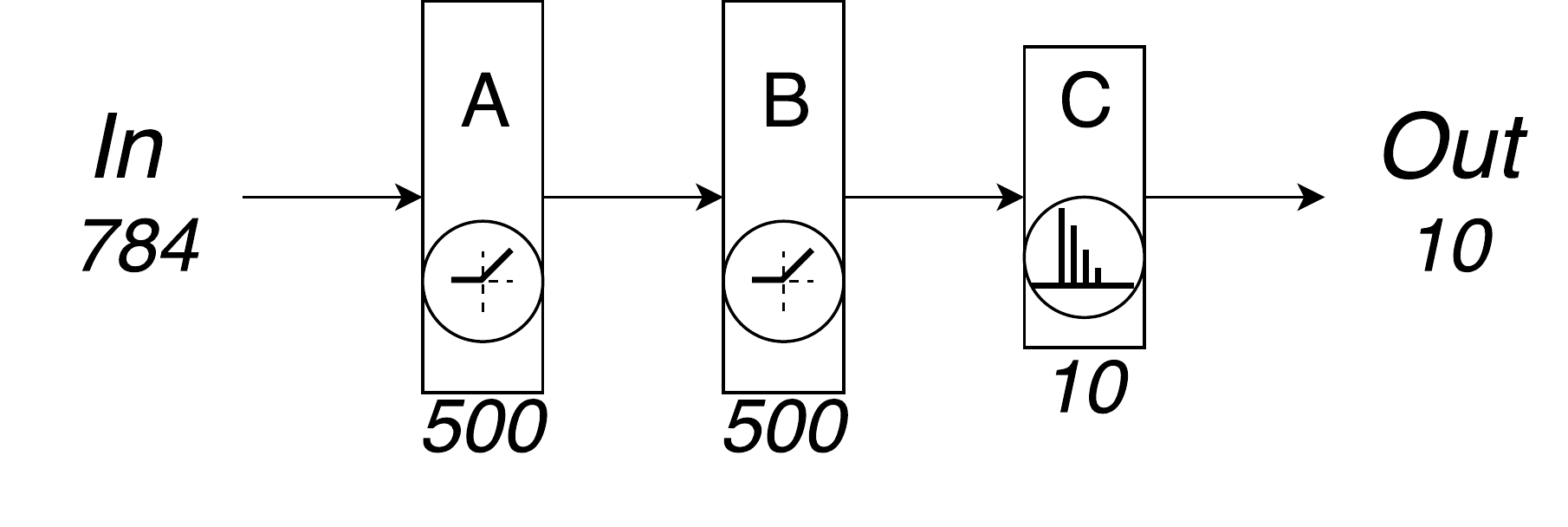}\label{fig:sc_topology}} \hfill
    \subfloat[Generative adversarial network (GAN). Its \name{} representation comprises 3 \Network{}s and is shown in Figure \ref{fig:GAN-individual-nets}.]{\includegraphics[width=.65\textwidth]{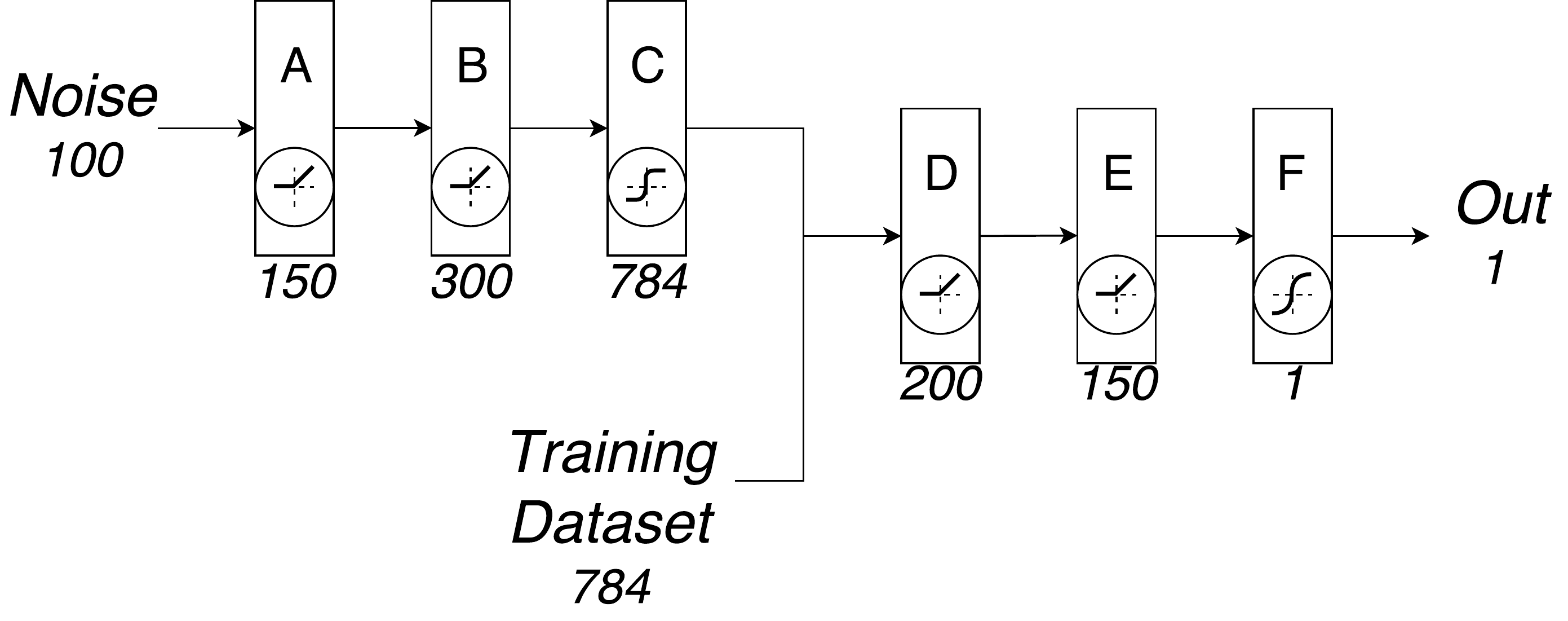}\label{fig:gan_topology}}

    \subfloat[Adversarial Autoencoder (AAE). Its \name{} representation comprises 7 \Network{}s and is shown in Figure \ref{fig:AAE-individual-nets}.]{\includegraphics[width=.84\textwidth]{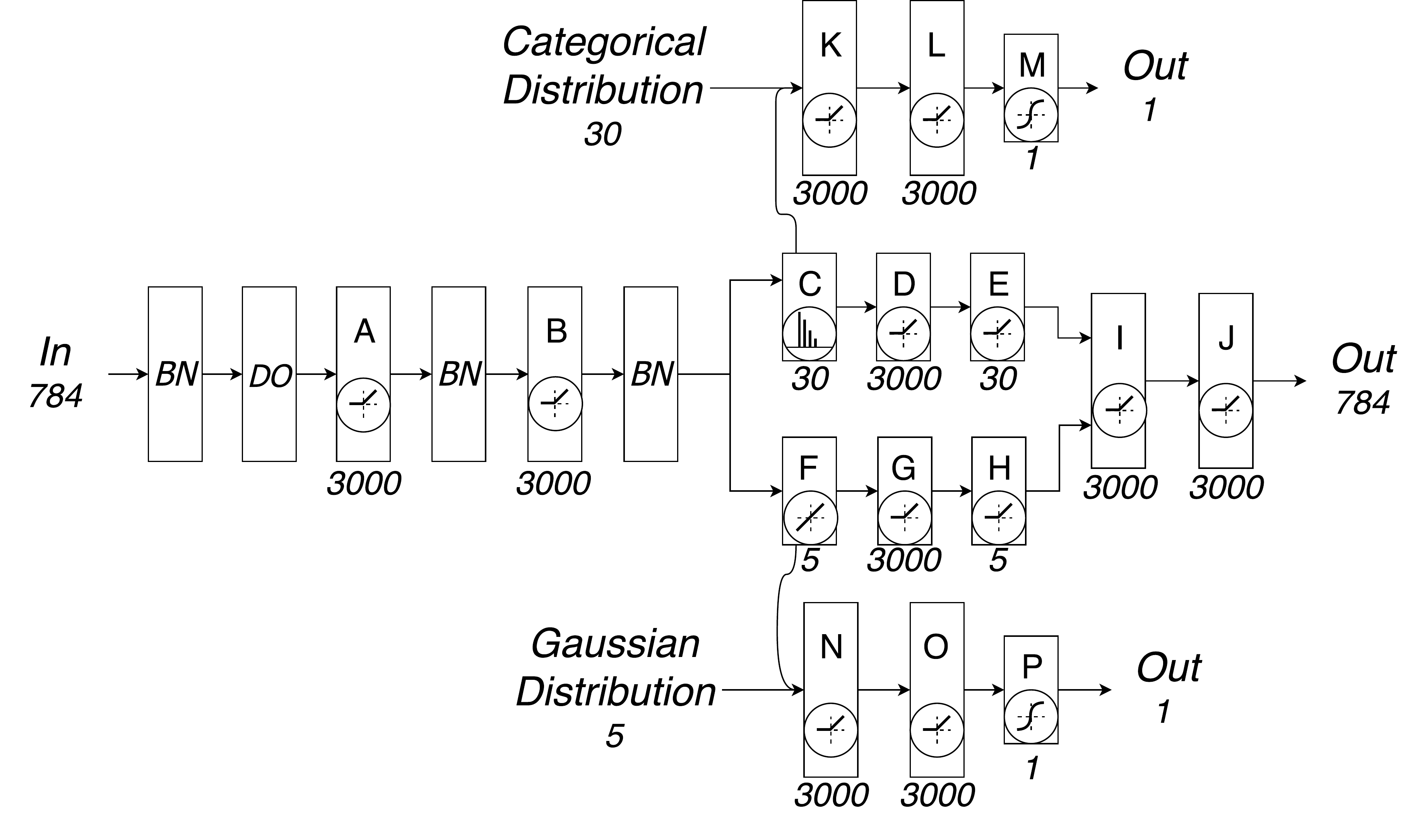}\label{fig:aae_topology}}

    \label{fig:topologies}
\end{figure}

\subsection{Simple classification network}
This use-case demonstrates \name{} in a single-network configuration. Algorithms and diagrams are omitted for brevity. Emphasis is put on the more involved use-cases of Sections \ref{sec:gan} and \ref{sec:aae}.

\paragraph{Topology:}
The simple classification network consists of one input layer, three stacked hidden layers (two of which carry rectified linear (ReLU) activations and one activated by SoftMax) and one output layer, as shown in Figure \ref{fig:sc_topology}. The layers are characterized by using the subclasses of \Layer{}: \InputLayer{}, \ocls{FullyConnectedLayer} and \OutputLayer{}. Activations are specified by the \ocls{hasActivationFunction} property. This ANN configuration is described by a single \Network{}, with data flowing from the input to the output layer. The use of input training or evaluation data is described by associating corresponding \DatasetPipe{}s with the \InputLayer{}.

\paragraph{Training:}
This use-case represents a single-objective, single-network configuration. Training it therefore involves calculating a single loss based on the network's output and updating all the weights so that this loss is minimized.
To is described by defining a \TrainingStrategy{} with a single \TrainingSession{}. The \TrainingSession{} performs a single \TrainingStep{} in every iteration. A \ocls{TrainingOptimizer} is associated with the \TrainingStep{} individual. The network's \ObjectiveFunction{} is described in terms of a \ocls{CostFunction} and is linked to the \Network{} individual. 

\paragraph{Evaluation:}
Classification effectiveness is measured in terms of the accuracy of predicted labels compared to the ground truth. We define a \DatasetPipe{} individual to join the \InputLayer{} with an evaluation dataset. An \ocls{Accuracy}:\Metric{}:\Function{} individual describes the evaluation metric. This metric and the evaluation dataset are associated with the configuration's \NetworkEvaluation{}.

\subsection{Generative Adversarial Network (GAN)}
\label{sec:gan}
GANs are multi-network configurations (Figure \ref{fig:gan_topology}).

\begin{figure}[t]
    \centering
    \caption{Individual \name{} \Network{}s for GAN. Boxes represent \Layer{}s.}
    \subfloat[Generator network]{\includegraphics[width=0.4\textwidth]{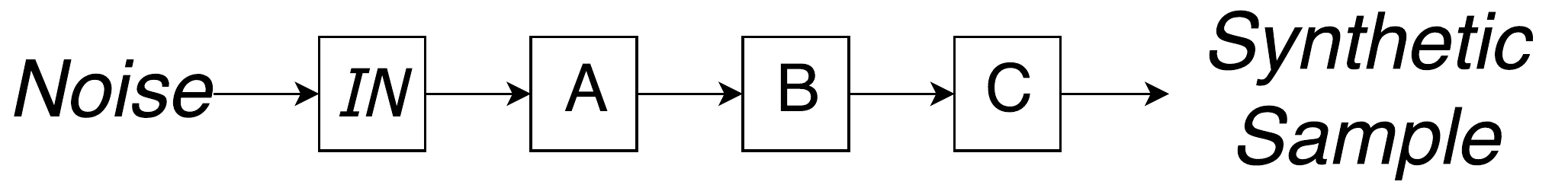}}\hfill
    \subfloat[Discriminator network]{\includegraphics[width=0.45\textwidth]{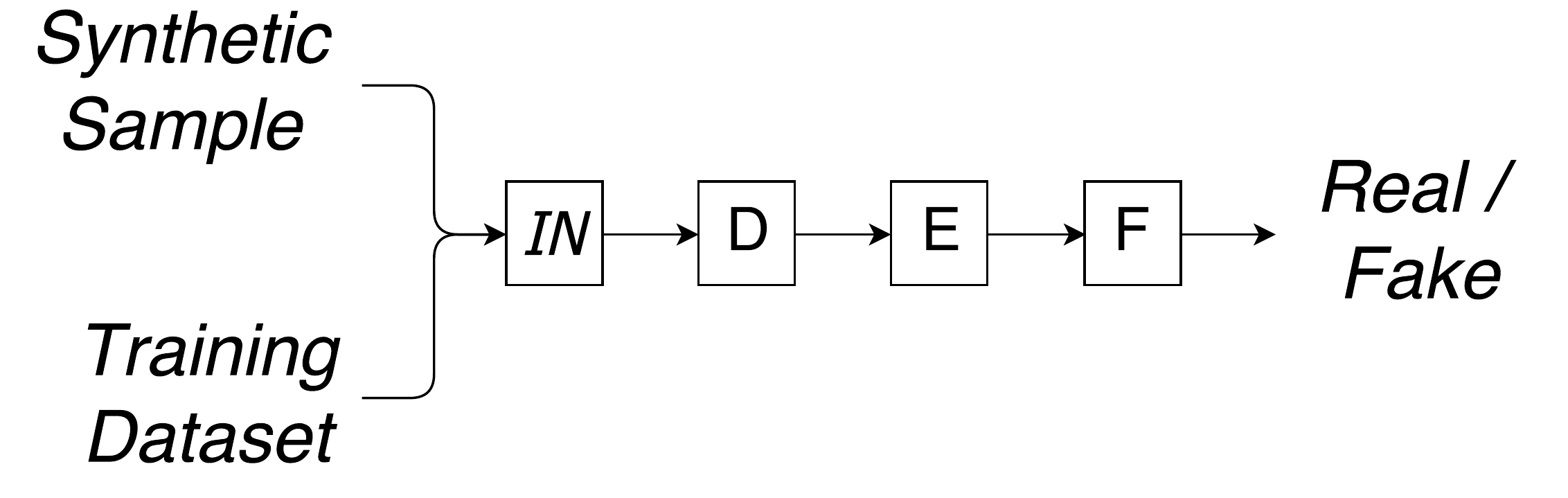}}
        
    \subfloat[Generative adversarial network]{\includegraphics[width=0.55\textwidth]{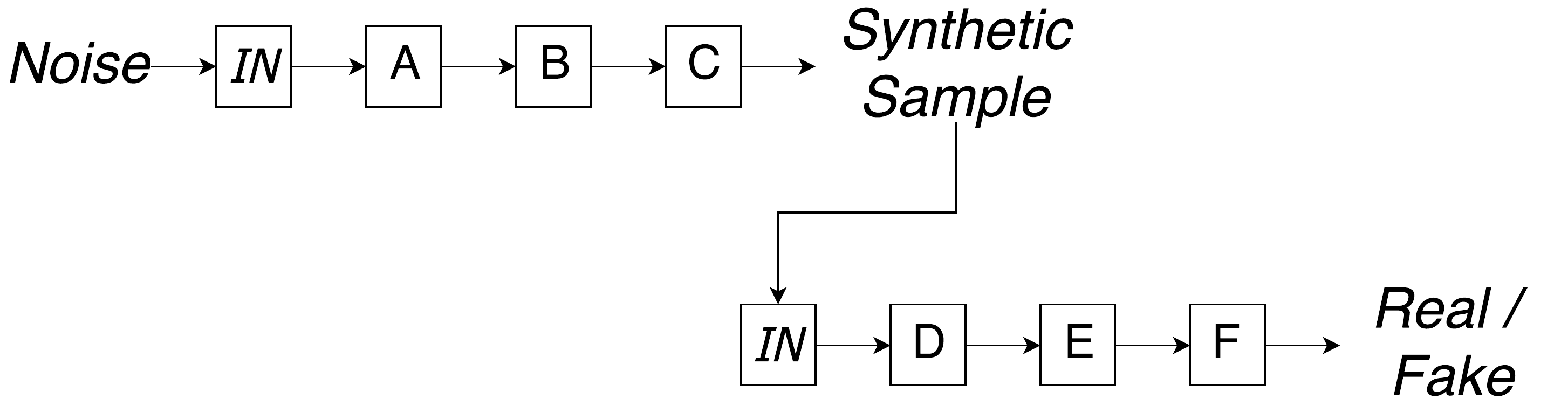}}
    \label{fig:GAN-individual-nets}
\end{figure}

\paragraph{Topology:}
GANs (Figure \ref{fig:gan_topology}) consist of two networks, the Generator and the Discriminator. Since in \name{} every network must have a single objective, a GAN is described in terms of three \Network{} individuals. These are named after the task that they perform \ie{} Generator, Discriminator and GAN. The Generator, Discriminator and GAN networks have task characterizations of \ocls{Generation}, \ocls{Discrimination} and \ocls{Adversarial} respectively. The GAN network describes the Adversarial process between the Generator and Discriminator and is composed of layers that are shared with the other two networks. Using the property \prop{sameLayerAs}, we are able to express layer sharing between these networks.  The individuals used to describe the overall topology are depicted in Figure \ref{fig:GAN-individual-nets}.

\paragraph{Training:}
GANs have multiple objectives, and \name{} associates each objective with a \Network{} individual. As the actual objective function is the same, it is associated with every network but with different \Labelset{} individuals. While the Generator and the Discriminator are trained alternately, we describe the case where the Discriminator gets updated 5 times per 1 Generator update -- such strategies often improve an adversarial network's learning efficiency. This procedure is described by a \TrainingLoop{} along with appropriate \ocls{TrainingSingleForwardOnly} and \ocls{TrainingSingle} individuals (Algorithm \ref{alg:gan-train}). Figure \ref{fig:GAN-train} shows the \name{} equivalent.

\begin{algorithm}
    \caption{Pseudocode for training an example GAN network. Comments point to relevant individuals in the \name{} model (Figure \ref{fig:GAN-train}).}
  \label{alg:gan-train}
\begin{verbatim}
DO UNTIL convergence:  # gan_session
   FOR i = 1..5:  # gan_trainloop
      train(Discriminator, train_data)  # gan_discriminate_mnist
      gen_output := compute_out(Generator, random_noise)  # gan_gen_fpass
      train(Discriminator, gen_output)  # gan_discriminate_generatorout
   END
   train(GAN, random_noise)  # gan_generatorstep
END
\end{verbatim}
\end{algorithm}

\paragraph{Evaluation:}
GANs are evaluated in terms of generated sample quality, given random noise as input. In \name{} we create a log-likelihood Parzen window estimate \Metric{} associated with a \NetworkEvaluation{}. GAN is evaluated on the Generator network output. Optionally, we can describe the training outcome as a \TrainedModel{} individual for storing purposes or further evaluation.

\subsection{Adversarial Autoencoder (AAE)}
\label{sec:aae}

Adversarial Autoencoders are multi-network configurations (Figure \ref{fig:aae_topology}). This example describes the clustering variant. 

\begin{figure}
    \centering
    \caption{Individual \name{} \Network{}s for AAE. Boxes represent \Layer{}s.}

    %

    %

    
    \subfloat[Autoencoder]{\includegraphics[width=0.65\textwidth]{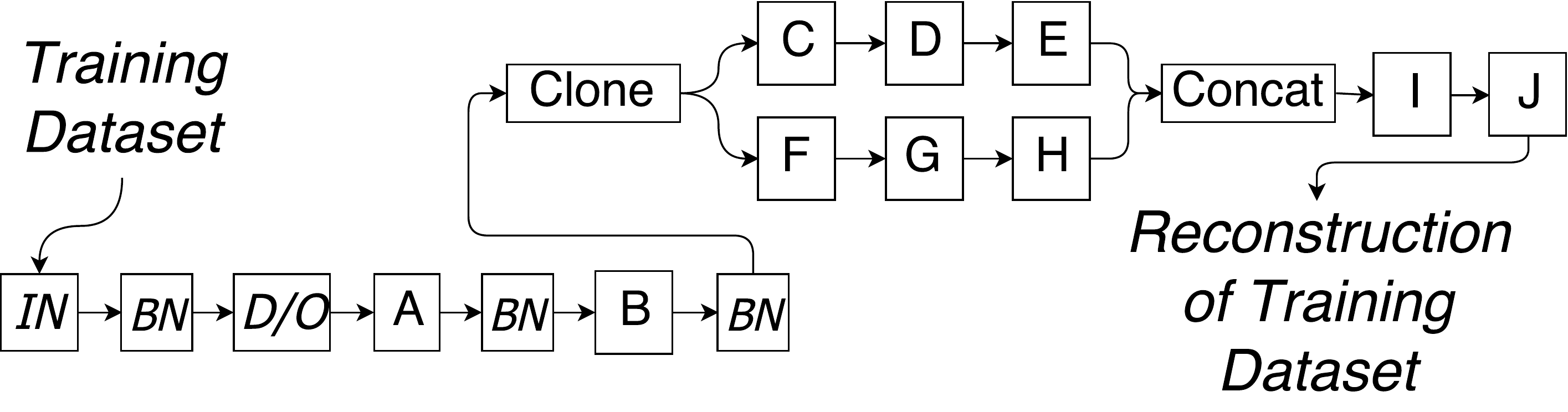}}

    \subfloat[Label discriminator]{\includegraphics[width=0.47\textwidth]{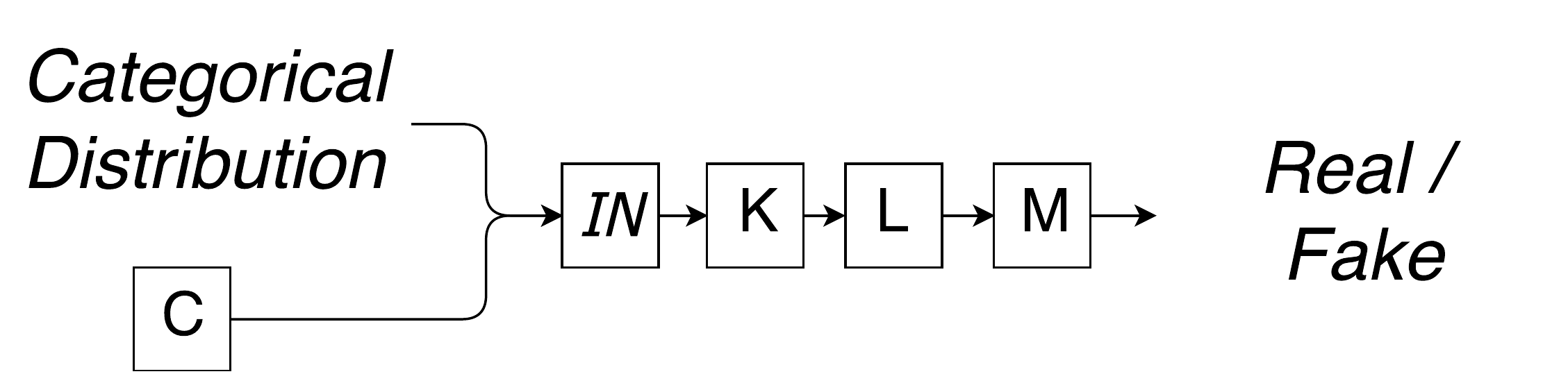}}
    \hfill
    \subfloat[Style discriminator]{\includegraphics[width=0.47\textwidth]{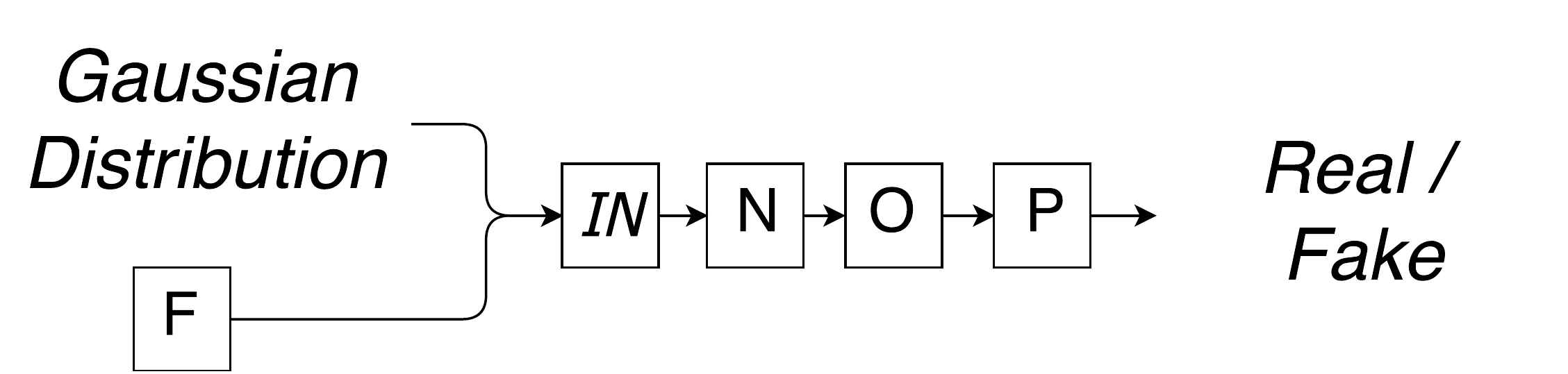}}
    
    \subfloat[Label generator]{\includegraphics[width=0.65\textwidth]{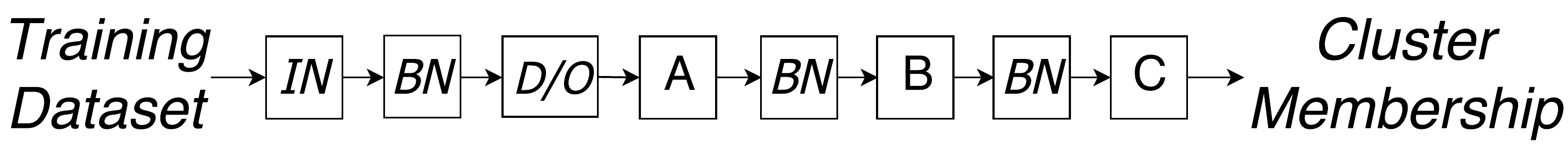}}

    \subfloat[Style generator]{\includegraphics[width=0.65\textwidth]{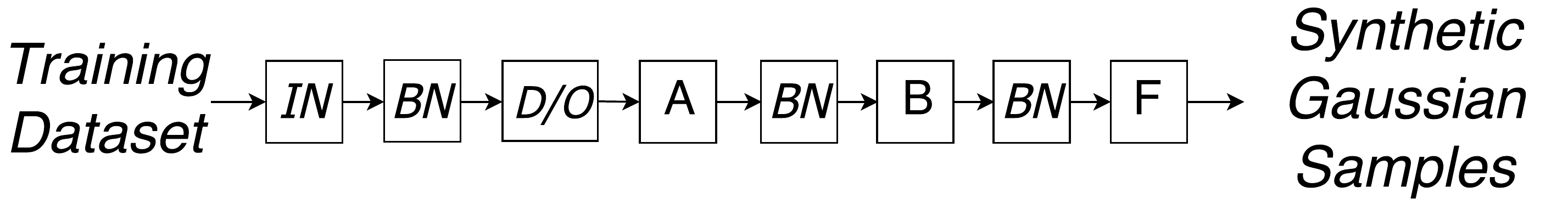}}
    
    \subfloat[Label generative adversarial network]{
        \includegraphics[width=0.6\textwidth]{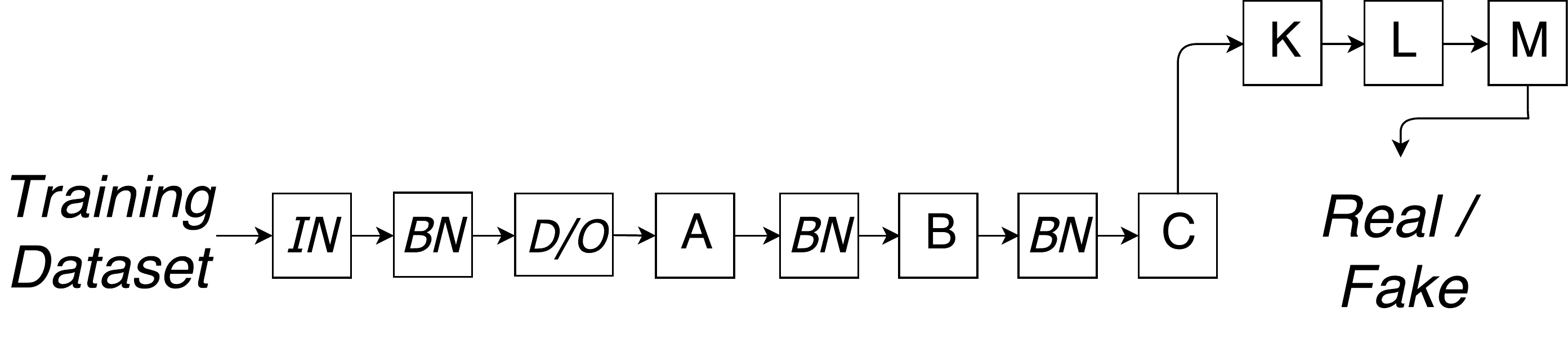}
    }

    \subfloat[Style generative adversarial network]{\includegraphics[width=0.6\textwidth]{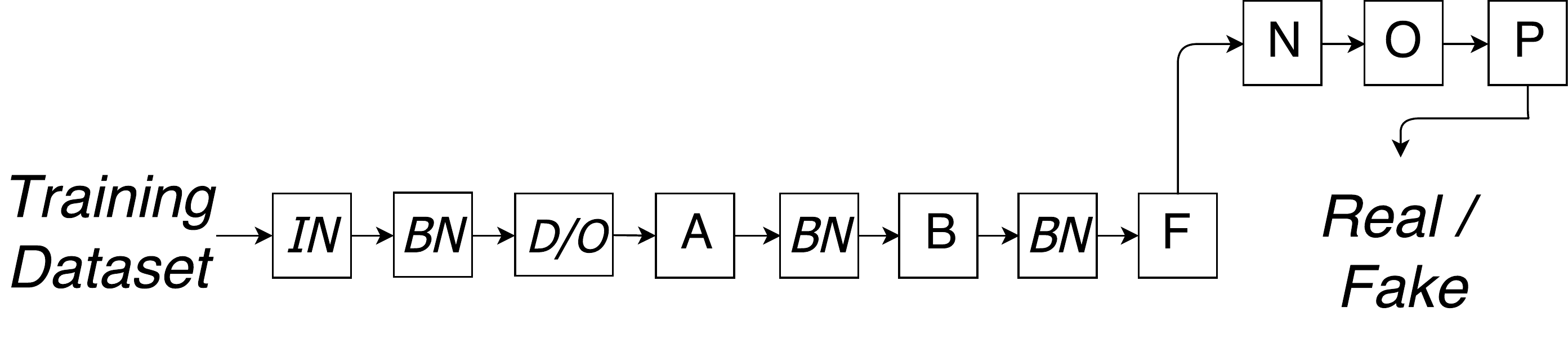}}
    \label{fig:AAE-individual-nets}
\end{figure}

\paragraph{Topology:}
Clustering in AAEs is a result of an adversarial process over the output of the network's encoder. \name{} can express this multi-network configuration using 7 \Network{} individuals, each of which performing a different task (Figure \ref{fig:AAE-individual-nets}). The encoder part of the Autoencoder is split into two branches (Style and Label) with these branches clustering the data sample and generating synthetic samples respectively. The splitting of the dataflow in the Encoder is expressed by a \SeparationLayer{} instance. Each of these two branches is also part of a Generator, as in an GAN. We describe the Discriminators using four \Network{} individuals, plus an additional one for the generative adversarial network as a whole. The decoder, which performs input reconstruction, concatenates each branch using an \AggregationLayer{}.

\begin{figure}[t]
    \centering
    \caption{Example training procedures for GAN and AAE configurations.}
    \subfloat[GAN training]{\includegraphics[width=0.46\textwidth]{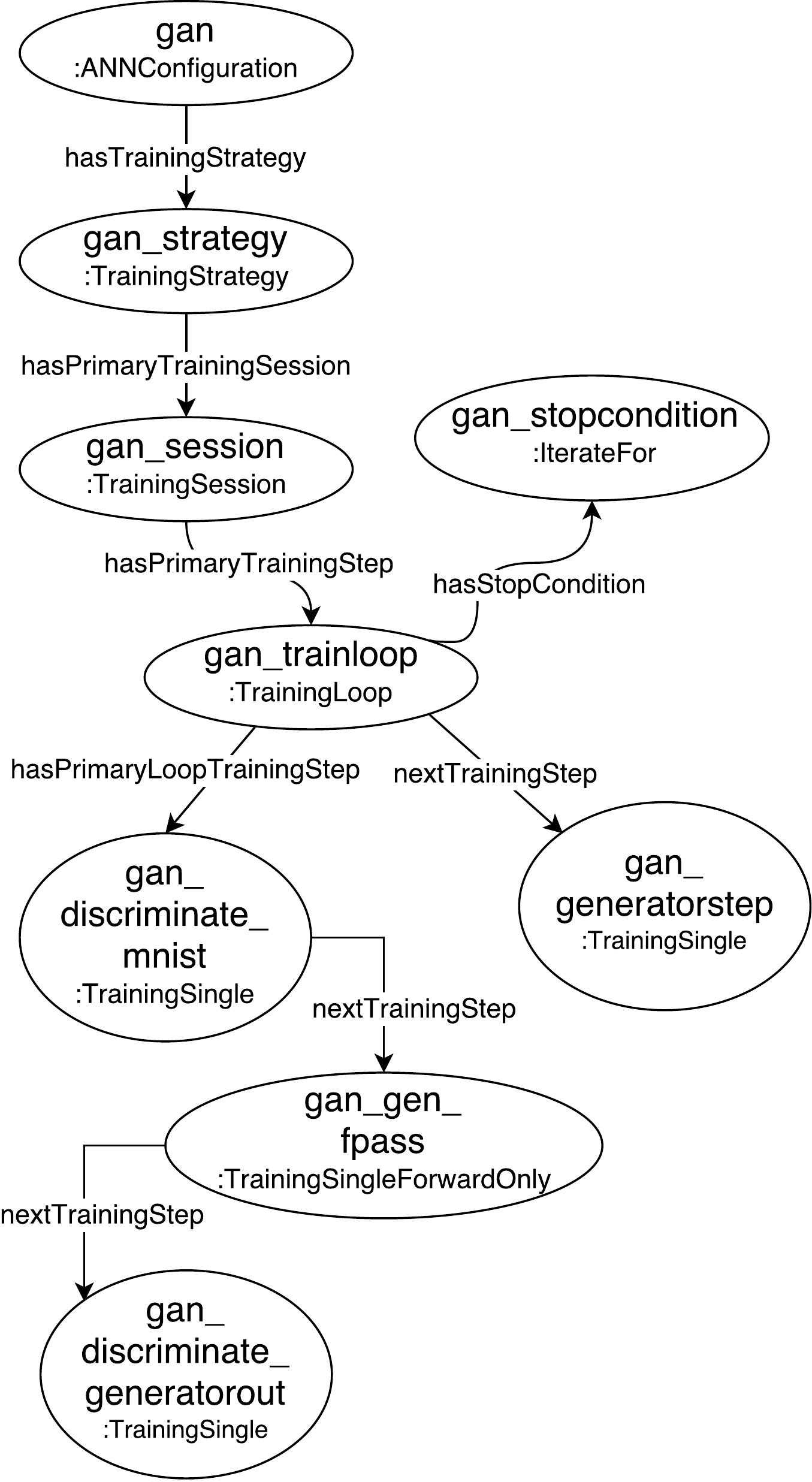}\label{fig:GAN-train}} \hfill
    \subfloat[AAE training]{\includegraphics[width=0.48\textwidth]{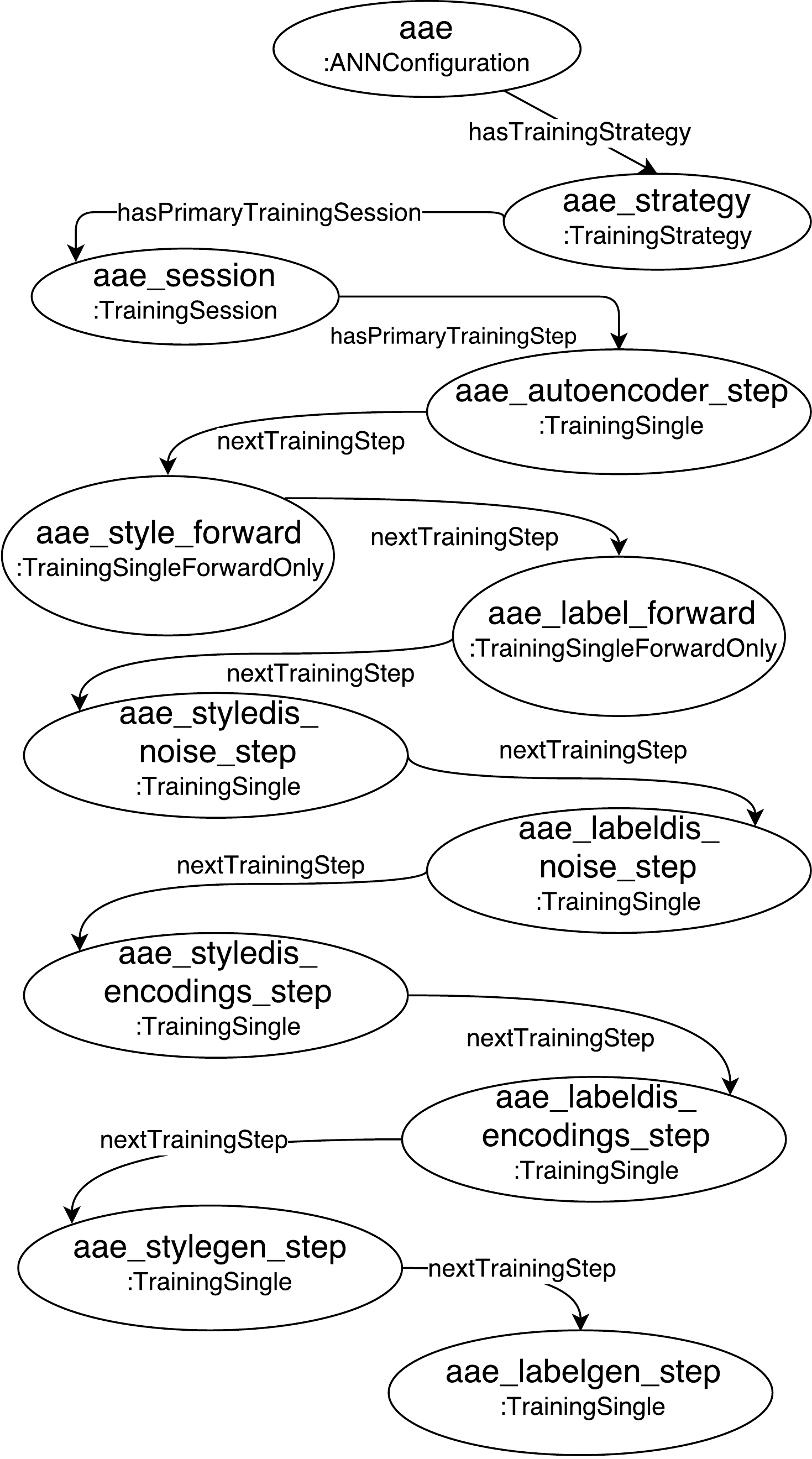}\label{fig:AAE-train}}
    \label{fig:GAN-AAE-train}
\end{figure}

\paragraph{Training:}
The training process involves multiple objectives that simultaneously update the shared layers. In AAEs the order in which we update the layers affects performance. We describe the training procedure via a sequence of \TrainingStep{} individuals. Since there are multiple shared layers in each network the property \prop{updatesLayer} is used to define which layers are updated in every \TrainingStep{}. Using \DatasetPipe{}s along with \TrainingStep{}s and the \prop{updatesLayer} property we can follow the dataflow in each individual network and observe the training process of the \ANNConfiguration{} as a whole. The training process is described in Algorithm \ref{alg:aae-train}, while  
Figure \ref{fig:AAE-train} shows the \name{} equivalent.

\begin{algorithm}
    \caption{Pseudocode for  training an example Adversarial Autoencoder. The \name{} model is shown in Figure \ref{fig:AAE-train}.}
  \label{alg:aae-train}
\begin{verbatim}
DO UNTIL convergence:  # aae_session
   train(Autoencoder, train_data)  # aae_autoencoder_step
   style := compute_out(StyleEncoder, train_data)  # aae_style_forward
   label := compute_out(LabelEncoder, train_data)  # aae_label_forward
   train(StyleDiscriminator, guassian)     # aae_styledis_noise_step
   train(LabelDiscriminator, categorical)  # aae_labeldis_noise_step
   train(StyleDiscriminator, style)  # aae_styledis_encodings_step
   train(LabelDiscriminator, label)  # aae_labeldis_encodings_step
   train(StyleGAN, train_data) # aae_stylegen_step
   train(LabelGAN, train_data) # aae_labelgen_step
END
\end{verbatim}
\end{algorithm}

\paragraph{Evaluation:}
Adversarial Autoencoders can be evaluated by measuring the classification accuracy on the predicted cluster membership. This can be modeled by creating an \ocls{Accuracy} metric and assigning it to a \NetworkEvaluation{}. Similar to the GAN example, performance is evaluated based on the output of a single network. The property \prop{basedOnTrainingStrategy} is used to associate a \NetworkEvaluation{} to a \TrainingStrategy{} individual.

\section{Query Examples}
\label{sec:queries}

The representation of deep learning configurations using \name{} constructs allows for the retrieval of ANN configurations or their constituents using various criteria. This allows for configurations or networks to be discovered based on their topological characteristics, their evaluation outcomes, or even procedural characteristics, \eg{} on features of the training process of a network. Such use-cases are highlighted by the following query examples.

\paragraph{Evaluation outcomes:} The following query retrieves training strategies based on their evaluation outcomes. Specifically, it seeks training strategies yielding evaluation scores greater than 0.7 in the classification task. Over the knowledge base accompanying the paper, the query retrieves the simple\_classification\_Strategy configuration:
\begin{Verbatim}[fontsize=\footnotesize]
select ?configuration ?evaluation_score where { 
    ?configuration a :ANNConfiguration.
    ?configuration :hasTrainingStrategy ?tstrategy;
                   :hasNetwork ?n.
    ?n :hasTaskType ?type.
    ?type a :Classification.
    ?evaluation a :NetworkEvaluation;
        :evaluatesNetwork ?n;
        :eval_score ?evaluation_score. {
        select ?tstrategy (count(?step) as ?steps) where {
            ?tstrategy :hasTrainingSession ?tsession.
            ?tsession :hasTrainingStep ?step
        } GROUP BY ?tstrategy HAVING (?steps > 2)
    } 
}
\end{Verbatim}

\paragraph{Topological characteristics:}
The following two queries search for ANN configurations having specific topological characteristics.
The next query retrieves the configurations comprising at least one network with a minimum of four hidden layers, of which at least one is a concatenation layer. In our example, it retrieves the AAE configuration.
\begin{Verbatim}[fontsize=\footnotesize]
select distinct ?c where {
    ?c a :ANNConfiguration;
       :hasNetwork ?n. {
        select ?n (count(?hl) as ?layers) where { 
        ?n :hasLayer ?l;
        :hasLayer ?hl.
        ?hl a :HiddenLayer.
        ?l a :ConcatLayer. 
        } GROUP BY ?n HAVING (?layers > 3)
    }
}
\end{Verbatim}

The following query retrieves configurations that include a network with at least one separation layer, whose branches lead to ReLU layers immediately before they merge again via concatenation. Executing the query over the demonstrative knowledge base returns the AAE\_AE  network.
\begin{Verbatim}[fontsize=\footnotesize]
select distinct ?n where { 
    ?n a :Network;
         :hasLayer ?l.
    ?l a :SeparationLayer. 

    ?l :nextLayer ?left;
       :nextLayer ?right.

    FILTER (?left != ?right)

    ?left :nextLayer+ ?c.
    ?right :nextLayer+ ?c.

    ?c a :ConcatLayer.
    ?c :previousLayer ?cpl.
    ?c :previousLayer ?cpr.

    ?cpl :hasActivationFunction ?fcpl.
    ?fcpl a :Relu.

    ?cpr :hasActivationFunction ?fcpr.
    ?fcpr a :Relu.    
}
\end{Verbatim}

\paragraph{Combined characteristics:} Last, the following query retrieves configurations comprising a clustering network that achieved an evaluation score greater than 0.5, with at least one training strategy involving at least one training session with more that two training steps. In our example, it retrieves the AAE configuration, which is associated with a training strategy satisfying the aforementioned constraints and an evaluation score equal to 0.68.

\begin{Verbatim}[fontsize=\footnotesize]
select ?configuration ?evaluation_score where { 
    ?configuration a :ANNConfiguration.
    ?configuration   :hasTrainingStrategy ?tstrategy;
                     :hasNetwork ?n.
    ?n :hasTaskType ?type.
    ?type a :Clustering.
    ?evaluation a :NetworkEvaluation;
                  :evaluatesNetwork ?n;
                  :eval_score ?evaluation_score. {
        select ?tstrategy (count(?step) as ?steps) where {
               ?tstrategy :hasTrainingSession ?tsession.
               ?tsession :hasTrainingStep ?step
        } GROUP BY ?tstrategy HAVING (?steps > 2)
    }     
}
\end{Verbatim}

\section{Conclusions and Future Work}
\label{sec:conclusions}
Increasingly complex deep learning configurations are being researched and used in a multitude of applications. These configurations require involved training procedures, while their effectiveness typically varies, depending on the task and the data size and type. This paper presented \name{}, an OWL ontology able to encode topological, training and evaluation characteristics of complex ANN configurations. Its purpose is to drive the development of knowledge bases capturing current and best practices of deep learning in order to enable researchers and practitioners understand existing systems and make better informed decisions when designing new ones. We have shown that \name{} is expressive enough to capture the topological and training characteristics of ANN configuration of high complexity using a small number of meaningful classes and properties. It also allows for the specification of sophisticated queries bridging different aspects of deep learning design. To the best of our knowledge this is the first ontology able to describe such a wide variety of ANN configurations.

We believe that efforts that contribute to the gathering and systematization of useful knowledge for R\&D in AI and data science can have tangible impact in these areas. However, for \name{} to reach its uptake and impact goals, software tools must be created to automate knowledge extraction from source implementations. Furthermore, aside from providing SPARQL endpoints, repositories and higher-level query services would be needed for users to make the most of the existing knowledge bases with minimal further training and therefore disruption to their work -- designing new deep learning algorithms. Last, even though deep learning is often and increasingly used on its own, it is also used in conjunction with other machine learning algorithms. On the front of modeling AI semantics, \name{} could be assimilated into future ontologies catering for the wider area of AI and machine learning. We anticipate that in the near future we will prototype accompanying tools and web services to allow users make the most of \name{}.

\bibliographystyle{splncs}
\bibliography{annett-o}

\end{document}